\documentclass[10pt,twocolumn,letterpaper]{article}

\usepackage{iccv}              

\usepackage{times}  
\usepackage{helvet}  
\usepackage{courier}  
\usepackage[hyphens]{url}  
\usepackage{graphicx} 
\usepackage{caption} 
\usepackage{algorithm}
\usepackage{algorithmic}

\usepackage{graphicx}
\usepackage{amsmath}
\usepackage{pifont}
\usepackage{multirow}
\usepackage[utf8]{inputenc} 
\usepackage[T1]{fontenc}    
\usepackage{url}            
\usepackage{amsfonts}       
\usepackage{nicefrac}       
\usepackage{microtype}      
\usepackage{xcolor}         
\usepackage{booktabs}

\definecolor{iccvblue}{rgb}{0.21,0.49,0.74}
\usepackage[pagebackref,breaklinks,colorlinks,allcolors=iccvblue]{hyperref}

\def\paperID{751} 
\def\confName{ICCV}
\def\confYear{2025}

\title{Leveraging BEV Paradigm for Ground-to-Aerial Image Synthesis}
\author{Junyan Ye\textsuperscript{\rm 1,2}\thanks{Equal contribution.},
    Jun He\textsuperscript{\rm 1*},
    Weijia Li\textsuperscript{\rm 1}\thanks{Corresponding author.}, \\
    Zhutao Lv\textsuperscript{\rm 1},
    Yi Lin\textsuperscript{\rm 1},
    Jinhua Yu\textsuperscript{\rm 1},
    Haote Yang\textsuperscript{\rm 2},
    Conghui He\textsuperscript{\rm 2,3} \\
    \textsuperscript{\rm 1}Sun Yat-Sen University, 
    \textsuperscript{\rm 2}Shanghai AI Laboratory,
    \textsuperscript{\rm 3}Sensetime Research\\
}

\usepackage{amsmath}
\usepackage{amssymb}
\usepackage{mathtools}
\usepackage{amsthm}
\usepackage{algorithm}
\usepackage{algorithmic}

\usepackage{multirow}       
\usepackage{amsmath}        
\usepackage{graphicx}       
\usepackage{caption}        
\usepackage{booktabs}       
\usepackage{adjustbox}      

\PassOptionsToPackage{table}{xcolor}

\usepackage{microtype}
\usepackage{graphicx}
\usepackage{booktabs} 
\usepackage{multirow}
\usepackage{makecell}

\usepackage{amsmath, amssymb, amscd, amsthm, amsfonts}
\usepackage{bm}
\usepackage{graphicx} 

\usepackage{amsmath,epstopdf,amssymb,color,url,multirow,multicol,verbatim,threeparttable,diagbox,makecell,booktabs}
\definecolor{light-gray}{gray}{0.82}
\definecolor{aliceblue}{rgb}{0.94,0.97,1.0}

\begin{document}

\maketitle

\begin{abstract}

Ground-to-aerial image synthesis focuses on generating realistic aerial images from corresponding ground street view images while maintaining consistent content layout, simulating a top-down view. The significant viewpoint difference leads to domain gaps between views, and dense urban scenes limit the visible range of street views, making this cross-view generation task particularly challenging. In this paper, we introduce SkyDiffusion, a novel cross-view generation method for synthesizing aerial images from street view images, utilizing a diffusion model and the Bird’s-Eye View (BEV) paradigm. The Curved-BEV method in SkyDiffusion converts street-view images into a BEV perspective, effectively bridging the domain gap, and employs a "multi-to-one" mapping strategy to address occlusion issues in dense urban scenes. Next, SkyDiffusion designed a BEV-guided diffusion model to generate content-consistent and realistic aerial images. Additionally, we introduce a novel dataset, Ground2Aerial-3, designed for diverse ground-to-aerial image synthesis applications, including disaster scene aerial synthesis, low-altitude UAV image synthesis, and historical high-resolution satellite image synthesis tasks. Experimental results demonstrate that SkyDiffusion outperforms state-of-the-art methods on cross-view datasets across natural (CVUSA), suburban (CVACT), urban (VIGOR-Chicago), and various application scenarios (G2A-3), achieving realistic and content-consistent aerial image generation. The code, datasets and more information of this work can be found at \url{https://opendatalab.github.io/skydiffusion/}.

\end{abstract}

\vspace{-5mm}
\section{Introduction}\label{sec:intro}

In fields such as land cover classification \cite{lv2023novel, tong2020land}, urban planning \cite{zhu2019understanding, zhao2019exploring, zhou2025urbench}, and disaster response \cite{kucharczyk2021remote, zhao2024see, wang2023hierarchical}, the analysis of aerial imagery plays a crucial role. The cross-view Ground-to-Aerial Image Synthesis task leverages ground street-view images to synthesize corresponding aerial images at the same location. Street-view images can be captured and uploaded by users on platforms such as Google Street View or Mapillary, offering high flexibility in data collection and frequent temporal coverage \cite{arrabi2024cross}. Therefore, Ground-to-Aerial Image Synthesis can support rapid response during natural disasters \cite{ou2023method} or fill historical gaps in high-resolution satellite imagery records \cite{cornebise2022open}, among other applications.

\begin{figure}[t]
    \centering
    \includegraphics[width=\linewidth]{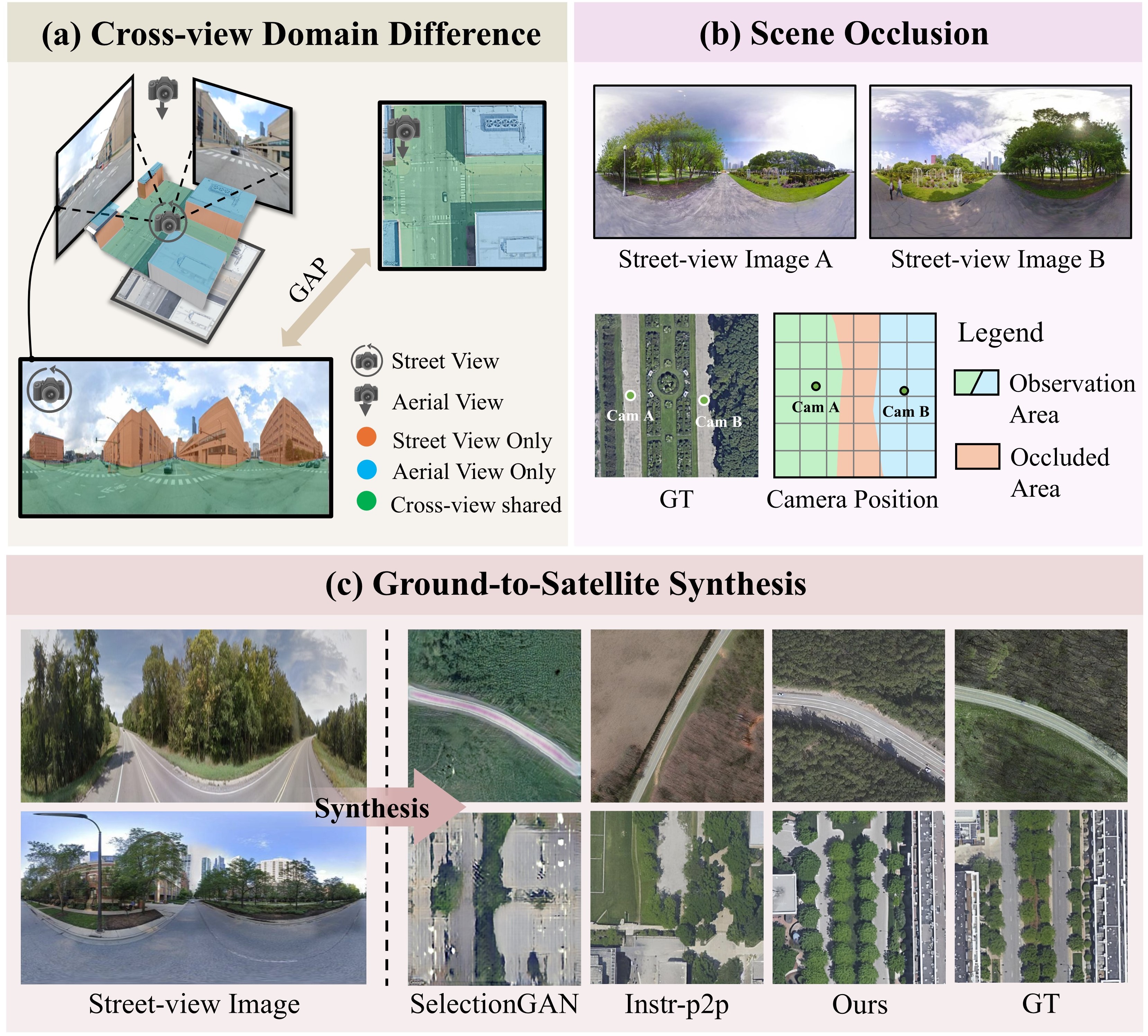}
    \caption{\textbf{Illustration of the cross-view image synthesis task.} (\textbf{a}) Challenges of cross-view domain gaps; (\textbf{b}) Challenges of occlusion in dense scenes; (\textbf{c}) Comparing our ground-to-aerial image synthesis method with existing cross-view synthesis methods.
    }
   \label{fig:task-illustration}
   \vspace{-5mm}
\end{figure}

Despite the potential of cross-view synthesis from ground-level to aerial imagery, significant challenges remain. The drastic viewpoint differences between street and satellite views lead to distinct object appearances, resulting in substantial domain gaps across image types\cite{lin2024geometry,li2023omnicity}. As shown in Figure \ref{fig:task-illustration} (a), in cross-view scenarios, street-view images primarily capture ground and building facade details, while satellite views present more rooftop and macroscopic summary information. Even with state-of-the-art diffusion methods, the domain gap between views often results in generated images that are realistic yet not true to reality; the synthesized satellite imagery may differ significantly from the actual content of corresponding street-views, as shown in Figure \ref{fig:task-illustration}(c).

Another challenge is object occlusion in dense urban settings, where tall buildings or trees limit the visibility range in street-views, which is substantially smaller than the coverage of satellite images \cite{li20243d}, as shown in Figure \ref{fig:task-illustration}(b). Aerial images synthesized from a single street-view image struggle to maintain consistency with real satellite images in unseen areas. An effective approach is to leverage multiple neighboring street-view images for satellite view synthesis. However, this poses a challenge for the model in effectively utilizing geographic location information from street-views and integrating features from multiple street-view images.

To address these challenges, this paper introduces SkyDiffusion, a ground-to-aerial image synthesis method based on diffusion models and the BEV paradigm, emulating a sky-view perspective. We first propose a Curved-BEV method, which transforms street-view images into Bird’s-Eye view to achieve domain alignment across viewpoints. Additionally, the Curved-BEV method introduces a multi-to-one mapping strategy, effectively integrating street-view images from different capture locations into a unified BEV space, thus overcoming the limited observation range in urban street-view images. Subsequently, we designed a BEV-guided diffusion model to generate realistic satellite images consistent with street-view content. Therefore, SkyDiffusion effectively overcomes the challenges of cross-view domain gaps and the limited observation range in densely occluded urban scenes, generating more realistic satellite images.

Moreover, existing cross-view datasets, such as CVUSA \cite{workman2015wide} and VIGOR \cite{zhu2021vigor}, were originally designed for cross-view retrieval tasks and thus offer limited diversity for cross-view synthesis. To address this, we introduce the Ground2Aerial-3 (G2A-3) dataset, which includes three novel aerial image generation scenarios: disaster emergency response, low-altitude UAV imagery, and historical high-resolution imagery generation tasks. The Ground2Aerial-3 dataset provides greater practical value and significance for exploring cross-view synthesis tasks in more diverse scenarios.
Our main contributions are as follows:

\begin{itemize}
\item We introduce SkyDiffusion, a novel ground-to-aerial synthesis method leveraging diffusion models and BEV paradigm to generate realistic, consistent aerial images.
\item We design a Curved-BEV method to transform street-view images into satellite views for domain alignment. It also includes "Multi-to-One" mapping strategy to enhance BEV perception range in densely occluded urban areas.
\item We introduce Ground2Aerial-3, a new ground-to-aerial image synthesis dataset, featuring disaster scene aerial image, low-altitude UAV image, and historical high-resolution satellite image 
synthesis tasks.
\item SkyDiffusion outperforms state-of-the-art methods on cross-view datasets across natural (CVUSA), suburban (CVACT), urban (VIGOR-Chicago), and various application scenarios (G2A-3), with an SSIM increase of 8.67\% and a FID reduction of 21.50\%.
\end{itemize}

\section{Related Work}\label{sec:relatedwork}

\textbf{Cross-view image synthesis} involves generating realistic scene images with significant viewpoint changes \cite{ze2025controllable,Sat2Density,li2024crossviewdiff,xu2024geospecific, zhang2023cross,ye2025satellite}. This task is common in computer vision and has broad applications in areas such as remote sensing image processing, virtual reality, and autonomous driving. It is mainly divided into aerial-to-ground and ground-to-aerial generation \cite{X-Seq, regmi2019cross}. In this field, synthesizing satellite images from street-views is a challenging yet important task. \citet{X-Seq} attempted to synthesize aerial views from a single street image, but the significant domain differences led to inconsistencies. \citet{tang2019multi} used street-view images and aerial semantic maps for image-to-image translation to synthesize target satellite images, but semantic maps are often unavailable in real scenarios. Additionally, the performance of GANs has also limited the quality of ground-to-aerial image synthesis in previous studies. Concurrent work \cite{arrabi2024cross} also synthesizes satellite images using text guidance and staged BEV semantic segmentation results, but these methods require additional text data and an extra semantic segmentation phase on street-view images. Unlike our method, it uses existing BEV methods and only supports one-to-one synthesis.

\begin{figure*}[!h]
    \centering
    \includegraphics[width=0.95\linewidth]{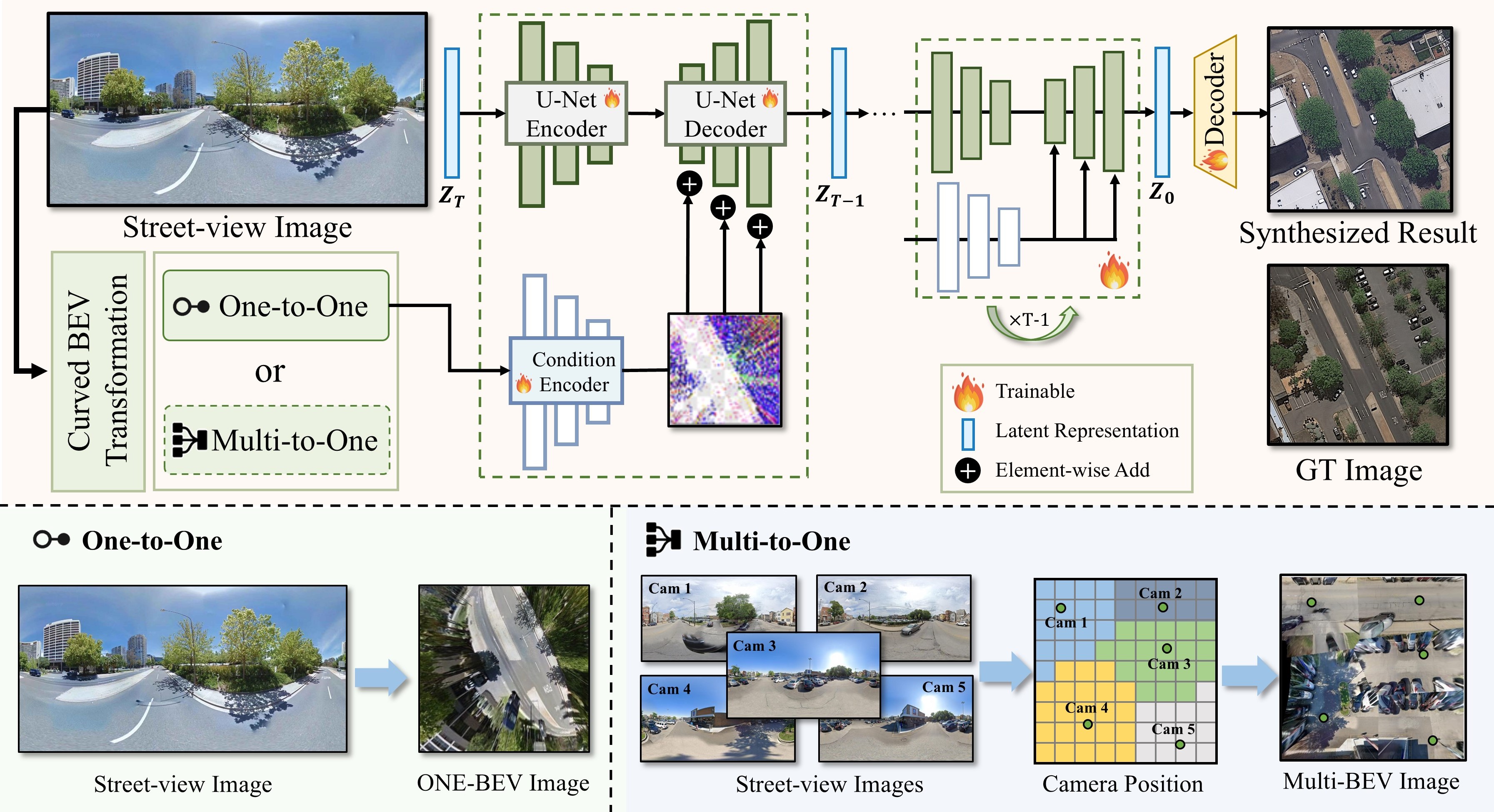}    
    \caption{\textbf{Overview of the proposed SkyDiffusion framework.} It include the curved BEV transformation and BEV-controlled diffusion model. The lower parts present the results of One-to-One and Multi-to-One BEV transformations, respectively.}
    \label{fig:pipline}
    \vspace{-4mm}
\end{figure*}

\noindent \textbf{BEV transformation} is a perspective domain conversion technique similar to polar transformation \cite{shi2019spatial,toker2021coming} and geo-transformation \cite{lu2020geometry}, widely used in retrieval \cite{ye2024cross}, navigation \cite{bevformer} and localization tasks  \cite{song2024learning,wang2024fine,shi2023boosting}. Traditional BEV transformation techniques, which rely on camera parameters \cite{sarlin2023orienternet} or depth estimation methods \cite{teng2024360bev, ye2024sg}, face challenges such as the difficulty in obtaining camera parameters and the high computational cost of depth estimation. Recently, some methods have been proposed to map street-view images to the BEV perspective using geometric relations at the image  \cite{wang2024fine} or feature level  \cite{shi2023boosting}. However, such methods are based on the ground plane assumption, where all mapped points are located below the camera, limiting the mapping of upper street scene features and causing information loss. Moreover, most existing panoramic BEV transformation methods primarily focus on one-to-one mappings, with relatively little research on multi-to-one BEV joint mappings.

\noindent \textbf{Novel view synthesis based on diffusion models.} Diffusion models have shown significant potential in recent studies on novel view synthesis  \cite{yan2025gpt,kawar2023imagic,gao2023magicdrive,ye2024loki,consis}. 
DreamBooth \cite{ruiz2023dreambooth} and AerialDiffusion \cite{AerialDiffusion} utilize text guidance to achieve top-down image generation by fine-tuning diffusion models. However, it is challenging to fully encapsulate the complexity of real-world scenes using textual information alone. Besides, diffusion-based image-to-image translation methods can also generate realistic satellite images \cite{li2023bbdm,insd,control}. However, the substantial domain gap between street-view images and satellite images presents a significant challenge in generating content-consistent satellite images.

\section{Methods}
As illustrated in Fig. \ref{fig:pipline}, this paper introduces SkyDiffusion, a novel method for synthesizing satellite images from corresponding street-view images. 
It initially applies a curved BEV transformation to the input street-view images, converting the perspective to a top-down view using either one-to-one or multi-to-one mappings (Section \ref{section3.1}). Subsequently, a BEV-guided diffusion model controls the synthesis of satellite images (Section \ref{section3.2}). Following this, we detail our network training process (Section \ref{section3.3}).

\subsection{Curved BEV Transformation}
\label{section3.1}

\textbf{One-to-One BEV.} 

Existing BEV methods based on ground plane assumptions (\(P(x,y,z=0)\)) and geometric relationships achieve the mapping from street-view to the BEV plane\cite{shi2023boosting,wang2024fine}. However, this assumption fails to map the content above the street-view image and results in noticeable distortion as points move away from the center area. Thus, we propose an enhanced curved BEV transformation to better capture cross-view information and address this issue.

\begin{figure}[t]
    \centering
    \includegraphics[width=0.9\linewidth]{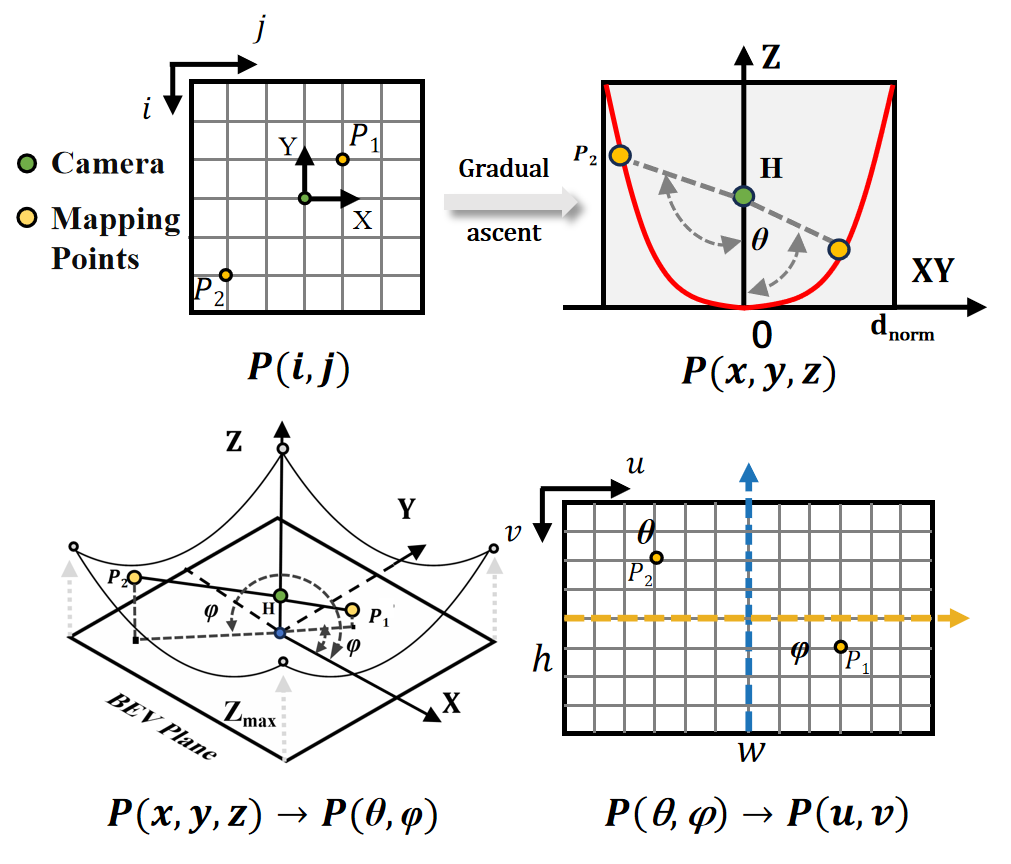}
    \caption{\textbf{Schematic of the curved BEV transformation.} It illustrates the mapping of two points on the BEV plane to the top and bottom of the street-view image during transformation.}
   \label{fig:cbev}
   \vspace{-5mm}
\end{figure}

As shown in Fig. \ref{fig:cbev}, we use \(P(i,j)\) to denote an index on the BEV space, and \(P(x,y,z)\) to illustrate a point in a three-dimensional coordinate system with the camera on the z-axis. We assume that the BEV plane is an upward curved surface, with the height (z-axis) increasing rapidly as the 3D points move away from the center (the central part of Fig \ref{fig:cbev}(a)). The following equation formulates the relationship between \(P(x,y,z)\) and \(P(i,j)\), where \( d_{norm} \) represents the normalized distance from the center, \( d_{max} \) is the maximum distance from the center, and \( \lambda \) is a scaling factor.

\vspace{-3mm}
\begin{gather}
\begin{cases} 
    x=j - \frac{l}{2}, \quad y=\frac{l}{2} - i\\
    z=d_{norm}^4 \times \lambda = \left(\frac{\sqrt{x^2 + y^2}}{d_{max}}\right)^4 \times \lambda
\end{cases} 
\label{eq:ij-xyz}
\end{gather}

Furthermore, we can convert \(P(x,y,z)\) to spherical coordinates \(P(\theta,\varphi)\). Based on the equirectangular projection properties of street-view images,  \(P(\theta,\varphi)\) corresponds to the panorama image index \( P(u,v) \). This allows us to establish the mapping relationship between \(P(x,y,z)\) and \( P(u,v) \) from the following equation, where \( H \) represents the camera height, and \( w \), \( h \) represent the width and height of the street image. More details are in the supplementary material.

\vspace{-3mm}
\begin{equation}
\left\{
\begin{aligned}
u &= \left[\text{arctan2}\left(y, x\right) + \pi\right] \frac{w}{2\pi} \\
v &= \left[\frac{\pi}{2} + \text{arctan2}\left(z - H, \sqrt{x^2 + y^2}\right)\right] \frac{h}{\pi}
\end{aligned}
\label{eq:xyz-uv}
\right.
\end{equation}

Following Eq. \ref{eq:ij-xyz} and Eq. \ref{eq:xyz-uv}, we achieve the projection from the street-view image to the bird's eye view without requiring camera calibration parameters or depth estimation methods \cite{teng2024360bev, ye2024sg}. This effectively maps upper street-view features like distant roads and buildings. Since the mapping relationship for each transformation is fixed, the computational overhead is minimal.

\noindent \textbf{Multi-to-One BEV.}
Due to the limitations imposed by dense buildings on street-view observations, the BEV sensing range is very limited in urban datasets like VIGOR. To address this, we propose a "Multi-to-one" BEV mapping method that unifies multiple street-view BEV mappings results \( \textit{BEV}_{\mathrm{cam}_k}\) into a single satellite coordinate system based on the relationships of camera position. We select the street-view mapping result closest to the capture point for overlapping areas following Eq.~\ref{eq:M-bev}.

\vspace{-3mm}
\begin{equation}
\left\{
\begin{aligned}
&\textit{BEV}\left(x, y\right) = \textit{BEV}_{\mathrm{cam}_k}\left(x - \Delta x_{k}, y - \Delta y_{k}\right) \\
&k = \arg \min_i \sqrt{(x - x_{\mathrm{cam}_i})^2 + (y - y_{\mathrm{cam}_i})^2}
\end{aligned}
\label{eq:M-bev}
\right.
\end{equation}

where \( \textit{BEV}\left(x, y\right) \) represents the unified BEV space corresponding to the satellite image, \( x, y\) corresponds to the index position of the BEV plane, \( \Delta x_{k} \) and \( \Delta y_{k} \) represent the offset of the street-view image from the satellite center, and \( x_{\mathrm{cam}_i} \) and \( y_{\mathrm{cam}_i} \) represent the coordinates of camera position in the satellite coordinate system. Additionally, during the implementation process, we also include boundary cropping as part of the image post-processing. 

\subsection{BEV-Controlled Diffusion Model}
\label{section3.2}

Image diffusion models progressively denoise images and generate samples from the training domain. The denoising process can occur in pixel space or in a latent space encoded from training data \cite{ddim,ddpm}.
In the forward process, diffusion models gradually add Gaussian noises to a ground truth image $x_0$ according to a predetermined schedule $\beta_1, \beta_2, \dots, \beta_T$:
\begin{equation}
q(x_t | x_{t-1}) = \mathcal{N}(x_t; \sqrt{1 - \beta_t}x_{t-1}, \beta_t I) 
\end{equation} 
where $x_t$ is a noised sample with noise level $t$. The reverse process involves a series of denoising steps, where noise is progressively removed by employing a neural network $\epsilon_\theta$ with parameters $\theta$. The reverse process involves a series of denoising steps, where noise is progressively removed by employing a neural network $\epsilon_\theta$ with parameters $\theta$. This neural network predicts the noise $\epsilon$ present in a noisy image $x_t$ at step $t$.

In the satellite image synthesis process, an effective feature embedding mechanism is required to ensure the generated satellite image aligns with the BEV image \(I_{bev}\). We use a conditional encoder to transform the BEV image into a latent space feature embedding, and then, similar to ControlNet \cite{control}, inject the features into the diffusion model using zero convolution layers and replicated SD modules. Specifically, we first use a small encoder to capture the cross-view associative information of the input BEV image \(I_{bev}\) such as road directions and building positions, to achieve BEV image feature embedding \(c_{bev}\). Notably, considering the potential feature distortions in BEV images, we employ a spatial attention mechanism to enhance feature capture while suppressing distorted features. 

\begin{equation}
c_{bev} = \mathcal{E}(I_{bev}).
\end{equation}

Subsequently, by using zero convolution layers and replicating the encoder and middle block structures and weights from the SD module, the extracted BEV features are injected into another SD module. Guided by the BEV image, the diffusion model incorporates cross-view associative information, synthesizing aerial views with consistent content.

\subsection{Network Training and Implementation Details}
\label{section3.3}
Given a set of cross-view image data, the image diffusion algorithm gradually adds noise to the satellite image, generating a noisy satellite image. It then learns a network $\epsilon_\theta$ to predict the noise added to the noisy satellite image $Z_t$ based on the given BEV image as the conditional input. Since text cannot fully describe the street-view scene, our cross-view image synthesis is essentially an image-conditioned synthesis task, with text not being our optimization target. This helps the model focus on generating satellite images based on the BEV conditions.

\vspace{-3mm}

\begin{equation}
\mathcal{L} = \mathbb{E}_{sat_0, t, c_{bev}, \epsilon \sim \mathcal{N}(0,1)} \left[ \left\| \epsilon - \epsilon_\theta (sat_t, t, c_{bev}) \right\|_2^2 \right]
\end{equation}

The loss $\mathcal{L}$ is the learning objective of the diffusion model, where $t$ represents the time step. We implement SkyDiffusion using the pre-trained Stable Diffusion v1.5 model~\cite{stable}, with an unlocked diffusion decoder and a classifier-free guidance scale of 9.0~\cite{cfg}. Final inference sampling uses 50 steps with the DDIM strategy~\cite{ddim}. Training is on eight NVIDIA A100 GPUs with a batch size of 128 for 100 epochs. 

\section{Dataset}
\label{sec:dataset}

\begin{figure}[t!]
    \centering
    \includegraphics[width=1\linewidth]{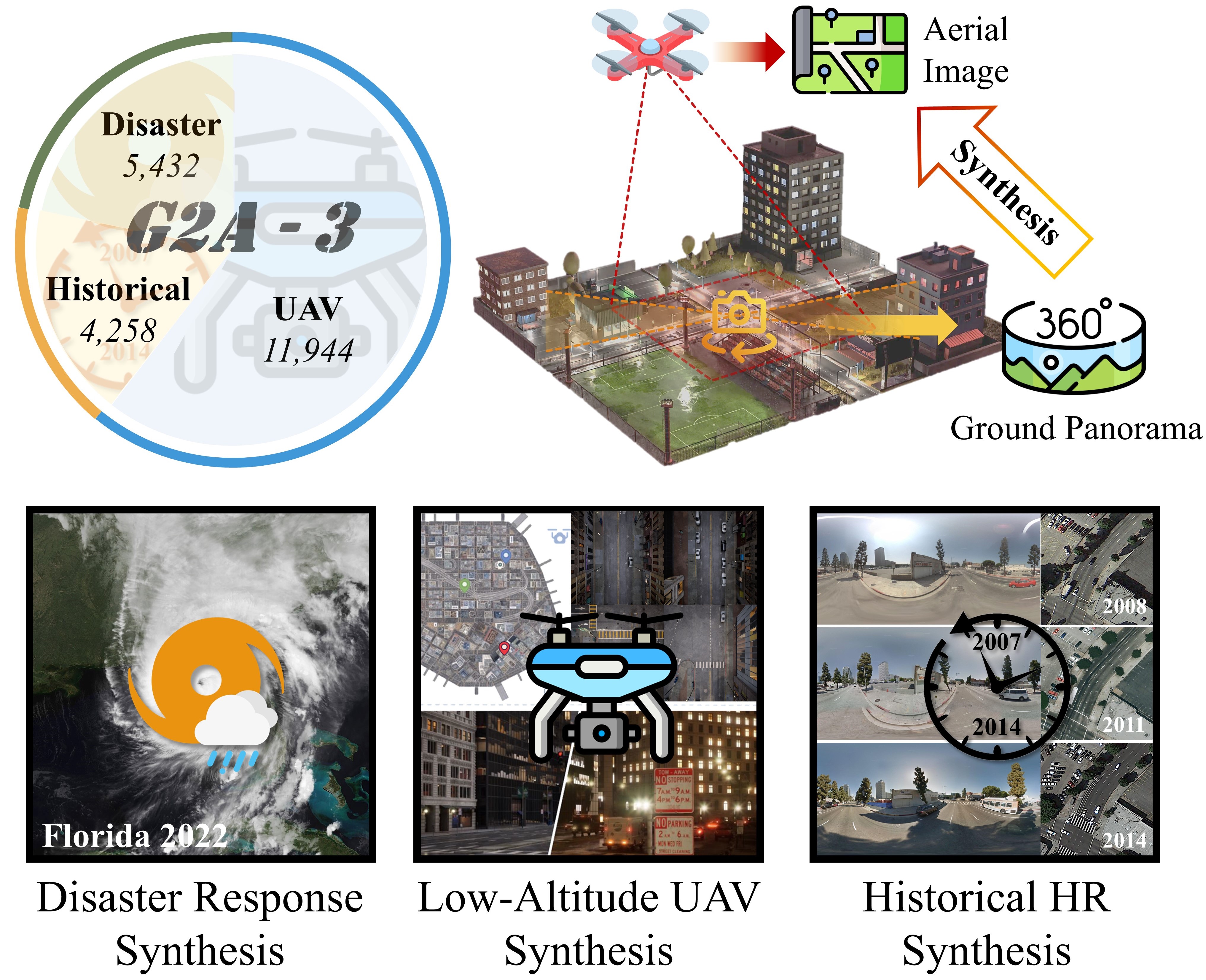}
    \caption{Illustration of the Ground2Aerial-3 dataset.}
    \label{fig:G2A-3}
    \vspace{-5mm}
\end{figure}

We propose Ground2Aerial-3, a multi-task cross-view synthesis dataset designed to explore the performance of cross-view synthesis methods in several novel scenarios. As shown in Figure \ref{fig:G2A-3}, G2A-3 contains around 20,000 street view and aerial images, covering disaster scene aerial image synthesis, low-altitude UAV image synthesis, and historical high-resolution satellite image synthesis. Due to the challenges in collecting data for specialized task scenarios, G2A-3 has gathered the available data to the greatest extent possible. The ground street-view images are 1024 $\times$ 512, with true north aligned at the center, and the aerial images are 512  $\times$ 512, aligned with the center of the ground images. The dataset for each task is divided into training and testing sets by region, with a 4:1 ratio.
The motivation and specific information for each task is as follows.

\subsection{Disaster Response Image Synthesis}

Satellite and aerial imagery are often limited by their capture frequency, making it difficult to obtain timely data for disaster response scenarios. Furthermore, acquiring multiple satellite images within a short time frame (e.g., 1 hour) to track disaster changes is typically infeasible. In such cases, synthesizing aerial images from street view images captured by ground cameras offers a fast and cost-effective solution, which can be further used for disaster assessment via comparative analysis with pre-disaster aerial imagery and assist in disaster geo-localization. CVIAN \cite{li2024cvdisaster} is a cross-view dataset containing street-view and high-resolution satellite imagery in Florida, USA, following Hurricane Ian in 2022. From CVIAN's street-view and satellite data, we filtered 2.7k image pairs of moderately or severely affected areas and adjusted the north direction of street-view images to the central column, preparing the data for the disaster scene aerial image generation task.

\subsection{Low-Altitude UAV Image Synthesis}

Low-altitude UAV imagery, with ultra-high resolution below 0.05 meters, has potential applications in tasks such as lane data collection for autonomous driving. However, the coverage and cost of UAV imagery are limited, and safety regulations further restrict coverage range and update frequency. Synthesizing low-altitude UAV images from more accessible street-view images presents a potential application scenario. Using virtual MatrixCity \cite{li2023matrixcity} data, we employed the UE engine to set up six single-view cameras at ground level, rendering and stitching them into panoramic street-view images. Specific details are provided in the supplementary materials. At the same locations, we positioned downward-facing cameras at an altitude of 20 meters to render low-altitude UAV imagery, creating a cross-view dataset for low-altitude UAV image synthesis tasks.

\subsection{Historical HR Satellite Image Synthesis}

Obtaining high-resolution (HR) satellite imagery, particularly ultra-high resolution images exceeding 0.3m, has been challenging. It wasn't until the recent launch of WorldView-3 (2014) and WorldView-4 (2018) that high-resolution coverage at near-global scales became available. Previously, ultra-high resolution images were often limited to small areas captured by aerial photography, with both spatial and temporal coverage constraints. Therefore, by utilizing historical street view images, we can provide information about past scenes to synthesize satellite or aerial images of the same location, offering additional supplementary information. We collected historical street view images from 2007 to 2014 for the Boston and Los Angeles areas on the Mapillary \footnote{\url{https://www.mapillary.com/}} and used corresponding aerial images from the massGIS platform \footnote{\url{https://www.mass.gov/info-details/massgis-data-layers}} as high-resolution satellite imagery for training and validation of synthesis results. Due to significant variations in the number of street-view images across different years, we still organized the dataset by region.

\begin{figure*}[h!]
    \centering
    \includegraphics[width=0.95\linewidth]{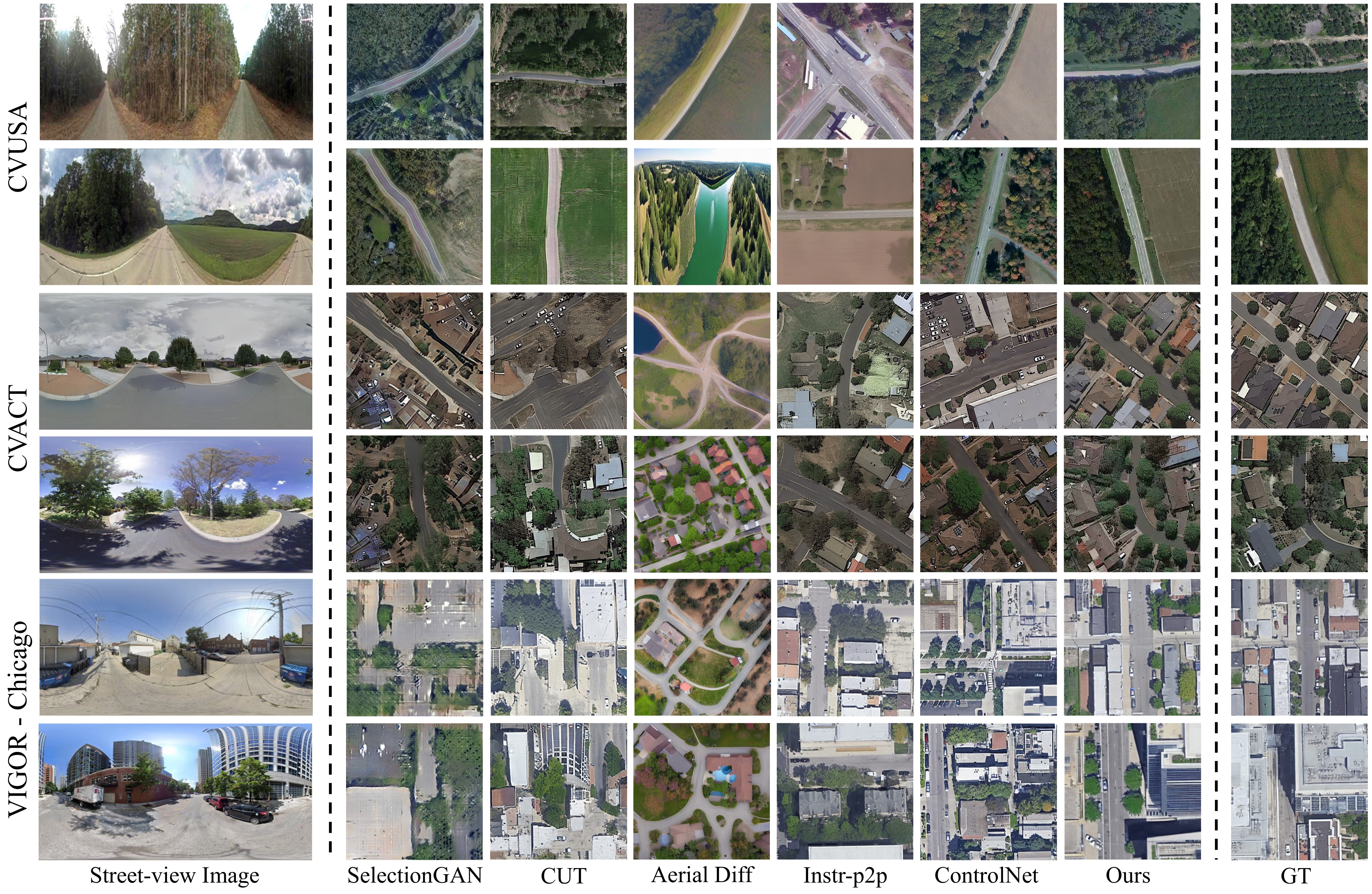}
    \vspace{-3mm}
    \caption{Qualitative comparison of different methods synthesis results on three datasets. }
   \label{fig:rural}
   \vspace{-5mm}
\end{figure*}

\section{Experiments}
\label{sec:Experiments}



In this section, we introduce the datasets utilized and the experimental setup. We then conduct both qualitative and quantitative comparisons of SkyDiffusion with state-of-the-art cross-view synthesis methods. Finally, we perform ablation studies to evaluate the effectiveness of each module.


%

\subsection{Experimental Setup}


\textbf{Dataset Details.}  CVUSA \cite{workman2015wide} and CVACT \cite{liu2019lending} are two commonly used cross-view datasets, containing one-to-one, center-aligned aerial and ground street-view images. They mainly covers suburban scenes, with low-rise buildings and open views. The CVACT dataset is regionally segmented. Although CVUSA uses random partitioning, its data distribution is sparse, with no overlap among satellite images.







VIGOR-Chicago \cite{zhu2021vigor} is a subset of the VIGOR cross-view dataset, containing the Chicago area. In this dataset, street-view panoramas and satellite images are not center-aligned, with multiple street-view images covering the same satellite image region, forming a multi-to-one mapping relationship. We modified the train-test split to a 4:1 ratio and ensured that the same satellite images do not appear in both the training and testing sets to prevent information leakage. The G2A-3 dataset has been described in Section \ref{sec:dataset}. 

\noindent \textbf{Parameter setting details.} All input street images are set to 512 $\times$ 1024 pixels, with the center column aligned to the north. Unlike previous methods that synthesize low-resolution satellite images (64 $\times$ 64), this experiment aims to meet practical needs by generating higher-resolution satellite images (512 $\times$ 512 pixels). For a multi-to-one mapping, to facilitate the merging of BEV mapping results from different street views, we set the hyperparameter $\lambda$ to 0, allowing for direct nearest-distance integration on the same horizontal plane. In this case, the BEV mapping of a single street-view image is similar to those in \cite{shi2023boosting,ye2024cross}, as it only needs to capture details near the capture point during the multi-to-one mapping process.

\noindent \textbf{Comparison Methods and Metric.} We compare our method with several state-of-the-art cross-view synthesis techniques, including GAN-based methods such as X-Seq \cite{X-Seq}, SelectionGAN (without semantic masks) \cite{tang2019multi}, CDTE \cite{toker2021coming}, and CUT \cite{park2020contrastive}, as well as diffusion-based new view synthesis and image transformation methods, including Aerial Diff \cite{AerialDiffusion}, GPG2A \cite{arrabi2024cross}, Instruct Pix2Pix (Instr-p2p) \cite{insd}, ControlNet \cite{control}, and img2img-turbo \cite{img2img-turbo}.
Except for the fine-tuned diffusion model, which requires brief textual descriptions, all other models use the same street-view input. 
Following previous work  \cite{lu2020geometry,toker2021coming,X-Seq}, we employ commonly used metrics such as SSIM, PSNR and LPIPS to evaluate the content consistency of the synthesized images, and use FID  \cite{FID} to assess the realism of images.

\begin{table*}[h]
    \centering
    \fontsize{9pt}{10pt}\selectfont
    \setlength{\tabcolsep}{0.35mm}
        \begin{tabular}{l|cccc|cccc|cccc}
            \toprule
            \multirow{2}{*}{\textbf{Method}} & \multicolumn{4}{c|}{\textbf{CVUSA}}  & \multicolumn{4}{c|}{\textbf{CVACT}}  & \multicolumn{4}{c}{\textbf{VIGOR-Chicago}}\\
            \cmidrule(lr){2-5} \cmidrule(lr){6-9} \cmidrule(lr){10-13}
            & \textbf{FID ($\downarrow$)} & \textbf{SSIM ($\uparrow$)} & \textbf{PSNR ($\uparrow$)} & \textbf{LPIPS ($\downarrow$)}  & \textbf{FID ($\downarrow$)} & \textbf{SSIM ($\uparrow$)} & \textbf{PSNR ($\uparrow$)} & \textbf{LPIPS ($\downarrow$)}  & \textbf{FID ($\downarrow$)} & \textbf{SSIM ($\uparrow$)} & \textbf{PSNR ($\uparrow$)} & \textbf{LPIPS ($\downarrow$)} \\
            \midrule
            X-Seq \cite{X-Seq}  & 161.16 & 0.084 & 11.97 & 0.706 & 190.12 & 0.042 & 12.41 & 0.661  & - & - & - & - \\
            CDTE \cite{toker2021coming}  & 122.84  & 0.143 & 10.01 & 0.694  & 160.81 & 0.091 & 12.59 & 0.663 & - & - & - & - \\
            SelGAN \cite{tang2019multi} & 116.57 & 0.129 & 12.38 & 0.742 & 100.21 & 0.116 & 12.04 & 0.684 & 149.53 & 0.127 & 11.77 & 0.778 \\
            Aerial Diff \cite{AerialDiffusion}& 136.18 & 0.103 & 10.06 & 0.855 & 127.29 & 0.108 & 10.24 & 0.878 & 123.16 & 0.141 & 11.49 & 0.831 \\
            CUT \cite{park2020contrastive} & 72.83 & 0.121 & 12.09 & 0.687  & 62.22 &  0.102 & 13.11  & 0.664 & 69.42 & 0.169 & \textbf{13.66} & 0.665 \\
            I2I-Turbo \cite{img2img-turbo} & 77.95 & 0.127 & 12.64 & 0.685 & 73.24 & 0.103 & 12.19 & 0.679 & 80.10 & 0.135 & 11.28 & 0.690 \\
            GPG2A \cite{arrabi2024cross} & 58.80 & 0.135 & 12.13 & 0.691  & 63.50 &  0.116 & 11.98   & 0.690 & 70.19 & 0.159 & 11.81 & 0.695\\
            Instr-p2p \cite{insd} & 38.01 & 0.138 & 12.29 & 0.697 & 49.62 & 0.116 & 12.51 & 0.682 & 50.30 & 0.163 & 10.62 & 0.689\\
            ControlNet \cite{control} & 32.45 & 0.149 & 12.63 & 0.650 & 62.21 & 0.115 & 11.95 & 0.682 & 53.27 & 0.170 & 10.38 & 0.666\\
            \midrule
            Ours  & \textbf{29.18} & \textbf{0.168} & \textbf{14.58}&  \textbf{0.635} & \textbf{36.48} & \textbf{0.118} & \textbf{12.85} & \textbf{0.645} & \textbf{45.29} & \textbf{0.186} & 11.69 & \textbf{0.661}\\
            \bottomrule
        \end{tabular}
    \caption{Quantitative comparison of different methods on CVUSA, CVACT and VIGOR-Chicago.}
    \label{tab:compariso_CVUSA_CVACT}
\end{table*}

\begin{figure*}[!t]
    \centering
    \includegraphics[width=0.98\linewidth]{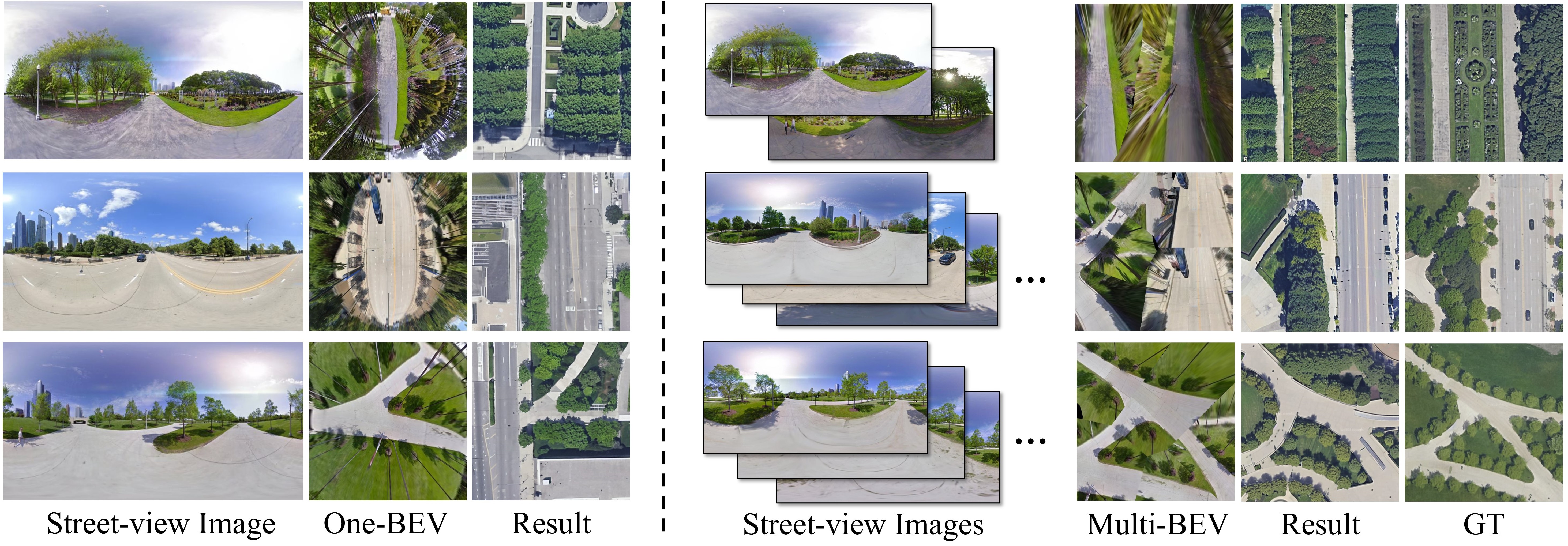}
    \vspace{-1mm}
    \caption{Qualitative comparison of synthesis results for One-to-One and Multi-to-one BEV.}
   \label{fig:multi2one}
   \vspace{-2mm}
\end{figure*}

\subsection{Comparison with State-of-the-Art Methods}

On the suburban CVUSA and CVACT datasets, SkyDiffusion achieved the outstanding results, as shown in Table \ref{tab:compariso_CVUSA_CVACT}. Compared to state-of-the-art methods, it reduced FID by 25.72\% and increased SSIM by 7.68\%, demonstrating its superiority in synthesizing realistic and consistent satellite images. The visual results in Fig. \ref{fig:rural} show that GAN-based cross-view methods generate excessive artifacts and blurriness. While diffusion-based methods such as Aerial Diff and Instr-p2p produce highly realistic aerial views, they lack content correlation with the street-view images. In contrast, our method benefits from BEV perspective transformation, enabling effective capture of cross-view associative information. SkyDiffusion generates realistic images with consistent content layout, including road direction, tree placement, and building distribution. Additional visual results can be found in the supplementary materials.

In the urban VIGOR-Chicago dataset, our method reduced FID by 14.98\% and improved SSIM by 9.41\% compared to the state-of-the-art method, as shown in Table \ref{tab:compariso_CVUSA_CVACT}. The VIGOR results for SkyDiffusion use only a single street-view image. The results using Multi-to-One BEV generation with multiple street-view images can be found in the ablation study. The visual results in Fig. \ref{fig:rural} show that our method maintains consistent associative information such as crosswalks and lanes. While the non-associative information on building rooftops cannot be consistently generated, it still produces reasonable and realistic results.

\begin{table}[!h]
    \centering
    \fontsize{9pt}{10pt}\selectfont
    \setlength{\tabcolsep}{0.8mm}
    \begin{tabular}{lcccc}
        \toprule
        \textbf{Method} & \textbf{FID ($\downarrow$)} & \textbf{SSIM ($\uparrow$)} & \textbf{PSNR ($\uparrow$)} 
 & \textbf{LPIPS ($\downarrow$)} \\
        \midrule
        \multicolumn{5}{l}{\textbf{CVACT}} \\
        Baseline & 62.21 & 0.115 & 11.95 & 0.682 \\
        Ours (BEV) & 42.84 & 0.117 & \textbf{13.62} & 0.673 \\
        Ours (C-BEV) & \textbf{36.48} & \textbf{0.118} & 12.85 & \textbf{0.645} \\
        \midrule
        \multicolumn{5}{l}{\textbf{VIGOR-Chicago}} \\
        Baseline & 53.27 & 0.170 & 10.38 & 0.666 \\
        Ours (BEV) & 48.63 & 0.175 & 11.01 & 0.687 \\
        Ours (C-BEV) & 45.29 & 0.186 & \textbf{11.69} & 0.661\\
        Ours (C-BEV Multi) & \textbf{31.90} & \textbf{0.205} & 11.15 & \textbf{0.651} \\
        \bottomrule
    \end{tabular}
    \caption{Ablation study of the Curved-BEV module. "Baseline" represents directly using street-view image,"BEV" and "C-BEV" denotes using standard BEV or Curved-BEV transformation, and "Multi" stands for Multi-to-One.}
    \vspace{-3mm}
    \label{tab:ablation}
\end{table}

\begin{table}[!t]
    \centering
    \fontsize{9pt}{10pt}\selectfont
    \setlength{\tabcolsep}{2mm}
    \begin{tabular}{lcccc}
        \toprule
        \textbf{Method} & \textbf{FID ($\downarrow$)} & \textbf{SSIM ($\uparrow$)} & \textbf{PSNR ($\uparrow$)} & \textbf{LPIPS ($\downarrow$)} \\
        \midrule
        \multicolumn{5}{l}{\textbf{Disaster Scene aerial image synthesis}} \\
        ControlNet & 138.11 & 0.183 & 12.95 & 0.690 \\
        Ours & \textbf{96.16} & \textbf{0.195} & \textbf{13.07}&  \textbf{0.685} \\
        \midrule
        \multicolumn{5}{l}{\textbf{Low-altitude UAV image synthesis}} \\
        ControlNet & 52.76 & 0.371 & 16.55 & 0.598 \\
        Ours & \textbf{32.53} & \textbf{0.452} & \textbf{16.86} & \textbf{0.390}\\
        \midrule
         \multicolumn{5}{l}{\textbf{Historical HR satellite image synthesis}} \\
        ControlNet & 36.48 & 0.117 & 12.85 & 0.665 \\
        Ours & \textbf{33.29} & \textbf{0.129} & \textbf{14.02} & \textbf{0.651}\\
        \bottomrule
    \end{tabular}
    \caption{Quantitative comparison of methods on G2A-3 Dataset.}
    \label{tab:G2A}
    \vspace{-5mm}
\end{table}

\subsection{Ablation Study}

We conducted ablation experiments on CVACT and VIGOR, with results shown in Table \ref{tab:ablation}. Compared to directly using street-view images as input, the Curved-BEV method improves performance across multiple metrics by transforming street-view images into satellite views for domain alignment. This indicates that the Curved-BEV method aids in synthesizing more content-consistent satellite images. Furthermore, the Multi-to-one method further improves metrics compared to the one-to-one mapping, demonstrating its effectiveness in dense urban scenes. 

As shown in the visual results in Figure \ref{fig:multi2one}, using multiple street-view images for multi-to-one BEV mapping effectively addresses occlusions from tall objects like trees and buildings. This method effectively enhances the BEV sensing range in urban environments, facilitating the generation of large-scale satellite scenes. Additional ablation study results can be found in the supplementary materials. We will include additional experiments in the supplementary materials, such as comparisons between Curve-BEV and the standard BEV, as well as experiments on the hyperparameter $\lambda$, among others.

\subsection{Evaluation results on G2A-3 dataset}

As shown in Table \ref{tab:G2A}, the tasks on the G2A-3 dataset present certain challenges; however, our method achieves significant performance improvements over the commonly used image-conditioned synthesis method, ControlNet. SkyDiffusion reduces the FID by an average of 25.81\% and increases the SSIM by an average of 12.88\%. ControlNet generates satellite images from street-view input, creating a significant domain gap. In contrast, our method uses Curve-BEV transformation to bridge the viewpoint domain gap, enabling more consistent satellite image generation.

\begin{figure}
    \centering
    \includegraphics[width=\linewidth]{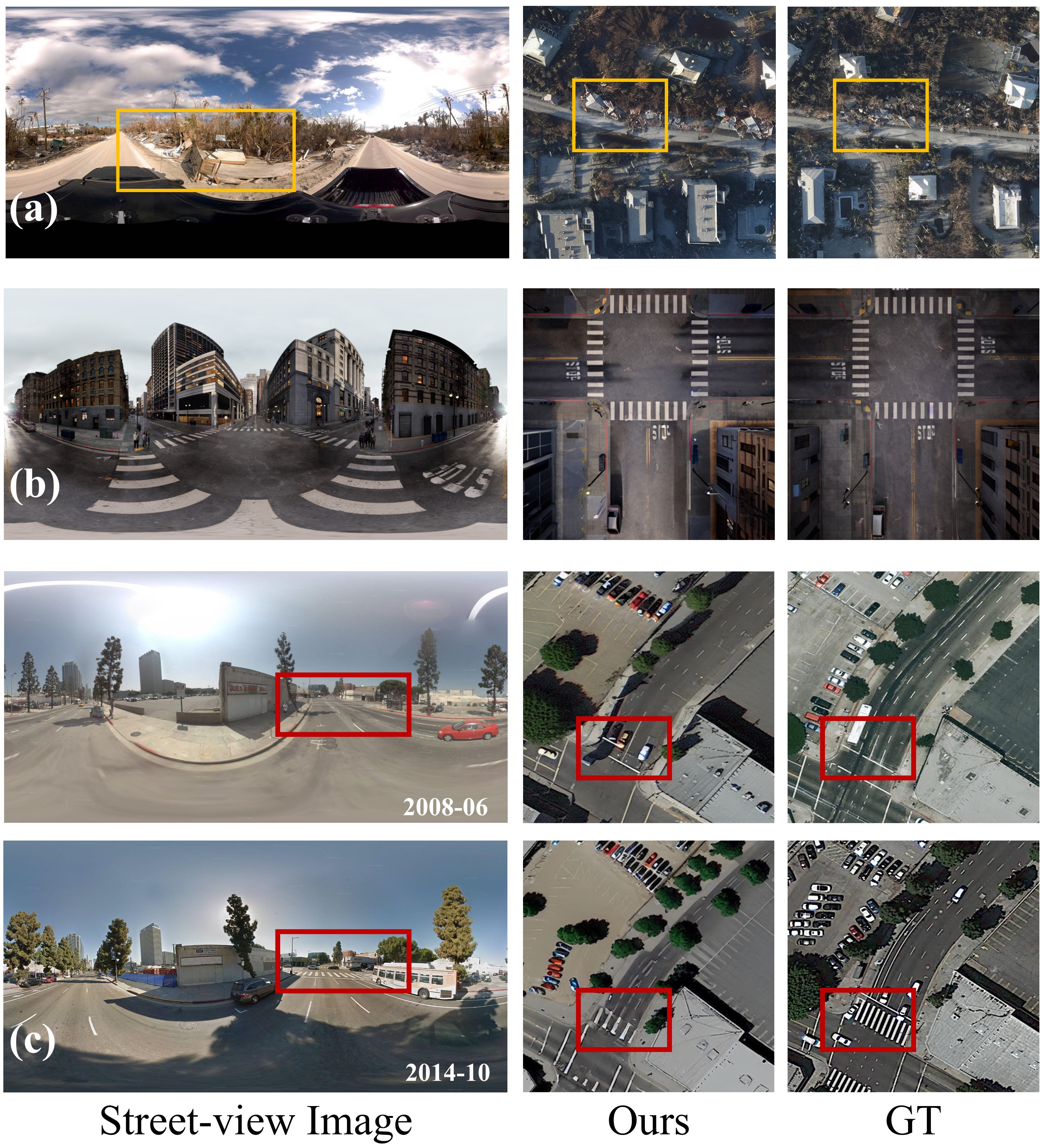}
    \caption{Qualitative Visualization Results on the G2A-3 Dataset.}
    \label{fig:G2A}
    \vspace{-5mm}
\end{figure}

\noindent \textbf{Disaster Scene Aerial image.} In Figure \ref{fig:G2A}(a), the key disaster-stricken regions, highlighted within the orange boxes, are distinctly visible in the generated satellite imagery. This provides a clear and comprehensive view of the disaster areas, with the potential to assist in subsequent assessment and in-depth analysis of the affected regions.

\noindent \textbf{Low-Altitude UAV Image.} The example in Figure \ref{fig:G2A}(b) demonstrates SkyDiffusion’s remarkable ability to synthesize low-altitude UAV images. Due to the lower altitude of UAV captures, there is greater overlap and closer domain similarity between scenes, resulting in superior synthesis quality. SkyDiffusion showcases outstanding aerial image synthesis capabilities, expanding its potential applications.

\noindent \textbf{Historical High-Resolution Satellite image.} Figure \ref{fig:G2A}(c) illustrates that between 2008 and 2014, the road underwent renovations, including the addition of crosswalks. SkyDiffusion successfully synthesizes realistic historical images from earlier periods, effectively addressing the absence of high-resolution Satellite imagery from those times. Additional visualizations on G2A-3 are available in the supplementary materials.

\vspace{-1mm}
\section{Conclusion}\label{sec:conclusion}
\vspace{-1mm}
In this study, we introduce SkyDiffusion, a novel approach specifically designed for street images to satellite images cross-view synthesis. SkyDiffusion operates solely with street images as input, utilizing a BEV Paradigm and diffusion models to generate satellite images. SkyDiffusion achieves state-of-the-art performance in both content consistency and image realism on across multiple cross-view datasets, demonstrating its superior capabilities. Additionally, we introduce a cross-view synthesis dataset, Ground2Aerial-3, featuring aerial image synthesis tasks for multiple new scenes, providing practical value and inspiration for future cross-view synthesis research.

\vspace{-1mm}
\section{Acknowledgement}
This work was supported in part by the Natural Science Foundation of Guangdong Province, China (Grant No. 2025A1515010400), the National Natural Science Foundation of China (Grant No. 42201358) and Shanghai Artificial Intelligence Laboratory.

{
    \small
    \bibliographystyle{ieeenat_fullname}
    \bibliography{main}
}

\end{document}